%% file: acl_latex.tex
\documentclass[11pt]{article}

\usepackage{acl}

\usepackage{times}
\usepackage{latexsym}
\usepackage{xspace}
\usepackage{xcolor}
\usepackage{pifont}
\usepackage{graphicx}
\usepackage{wrapfig}
\usepackage{caption}

\newcommand{\ours}{Always-Search Policy\xspace}
\newcommand{\policy}{Always-Search Policy\xspace}

\newcommand{\adaptive}{adaptive search\xspace}

\newcommand{\no}{\textcolor{red}{\ding{55}}}
\newcommand{\yes}{\textcolor{ForestGreen}{\ding{51}}}

\usepackage[T1]{fontenc}

\usepackage[utf8]{inputenc}

\usepackage{microtype}

\usepackage{inconsolata}

\usepackage{graphicx}
\usepackage{tabularx}
\usepackage{booktabs}
\usepackage{amssymb}
\usepackage{amsmath}
\usepackage{subcaption}
\usepackage{algorithm}
\usepackage{algpseudocode}
\usepackage{tcolorbox}
\tcbuselibrary{breakable} %
\tcbuselibrary{skins}     %
\usepackage{fontawesome5} %
\usepackage{hyperref}

\newtcolorbox{promptbox}[1]{
  colback=gray!5,   %
  colframe=gray!50, %
  coltitle=black,   %
  title={\textbf{#1}}, %
  fonttitle=\small\sffamily,
  fontupper=\small\ttfamily, %
  breakable,        %
  boxrule=0.8pt,    %
  left=2mm, right=2mm, top=2mm, bottom=2mm %
}

\usepackage[dvipsnames]{xcolor}

\title{Search, Do not Guess: Teaching Small Language Models \\to Be Effective Search Agents}

\definecolor{nyu-purple}{RGB}{112, 48, 160}  %
\definecolor{uiuc-blue}{RGB}{0, 32, 96}      %
\definecolor{ntu-blue}{RGB}{0, 112, 192}     %
\definecolor{equal-gray}{RGB}{166, 166, 166} %

\author{
  \textbf{Yizhou Liu}\textsuperscript{\textcolor{uiuc-blue}{U}\textcolor{equal-gray}{*}} \quad
  \textbf{Qi Sun}\textsuperscript{\textcolor{nyu-purple}{N}\textcolor{equal-gray}{*}} \quad
  \textbf{Yulin Chen}\textsuperscript{\textcolor{nyu-purple}{N}} \quad
  \textbf{Siyue Zhang}\textsuperscript{\textcolor{ntu-blue}{S}} \quad
  \textbf{Chen Zhao}\textsuperscript{\textcolor{nyu-purple}{N}} \\
  \textsuperscript{\textcolor{nyu-purple}{N}}New York University \quad
  \textsuperscript{\textcolor{uiuc-blue}{U}}University of Illinois Urbana-Champaign \\
  \textsuperscript{\textcolor{ntu-blue}{S}}Nanyang Technological University \\
  \faGithub \ \url{https://github.com/yizhou0409/Agentic-Rag} \\
  {\small \textcolor{equal-gray}{*}Equal contribution.}
}

\usepackage{tabularx}
\usepackage{array} %
\usepackage{booktabs}
\usepackage{float}
\usepackage{multirow}
\newcolumntype{L}{>{\raggedright\arraybackslash}X}

\begin{document}

\maketitle

\input{latex/0_abstract}
\input{latex/1_introduction}

\input{latex/2_taskformulation}

\input{latex/3_currentapproach}

\input{latex/4_search_only}
\input{latex/5_analysis}

\input{latex/7_relatedwork}

\input{latex/6_conclusion}

\bibliography{reference, latex/custom}

\appendix
\newpage
\input{latex/8_appendix}

\end{document}

%% file: latex/0_abstract.tex
\begin{abstract}
    
    Agents equipped with search tools have emerged as effective solutions for knowledge-intensive tasks. While Large Language Models (LLMs) exhibit strong reasoning capabilities, their high computational cost limits practical deployment for search agents. Consequently, recent work has focused on distilling agentic behaviors from LLMs into Small Language Models (SLMs). Through comprehensive evaluation on complex multi-hop reasoning tasks, we find that despite possessing less parametric knowledge, SLMs invoke search tools less frequently and are more prone to hallucinations. To address this issue, we propose \policy, a lightweight fine-tuning approach that explicitly trains SLMs to reliably retrieve and generate answers grounded in retrieved evidence. Compared to agent distillation from LLMs, our approach improves performance by 17.3 scores on Bamboogle and 15.3 scores on HotpotQA, achieving LLM-level results across benchmarks. Our further analysis reveals that adaptive search strategies in SLMs often degrade performance, highlighting the necessity of consistent search behavior for reliable reasoning.

\end{abstract}

%% file: latex/1_introduction.tex
\section{Introduction}

Language model based search agents iteratively issue search queries, reason over retrieved evidence, and adapt subsequent actions based on intermediate results. Such agents like Search-o1~\cite{li2025search} have shown strong potential for solving complex information-seeking problems. 

\begin{figure}[t!] 
    \centering
    \includegraphics[width=0.95\columnwidth]{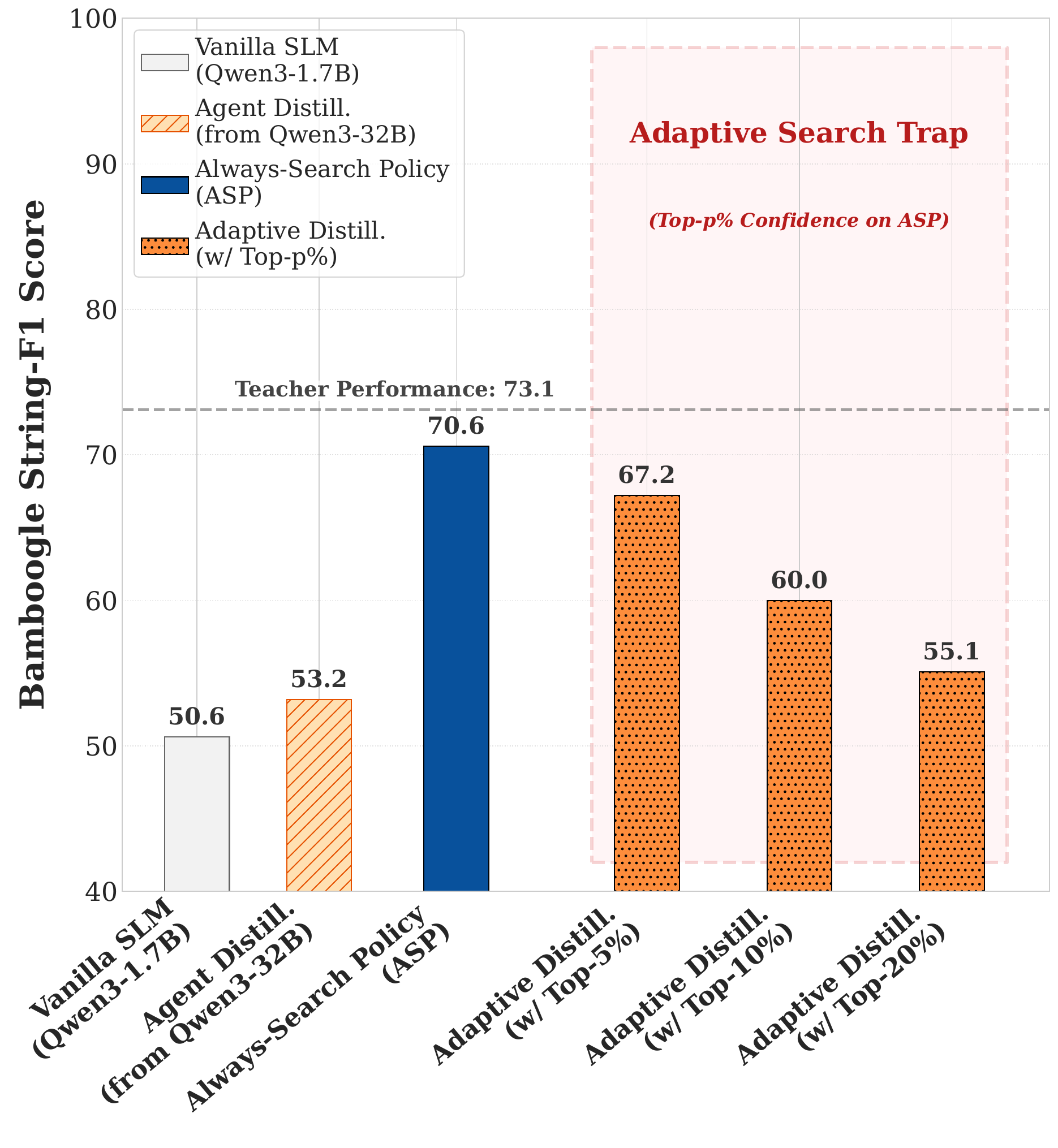}
    \caption{Left: With \ours, the distilled SLM significantly narrows the performance gap with the teacher model. Right: SLMs suffer from adaptive search and ASP is the most effective policy.} %
    \label{fig:bambooge_comaprison}
    \vspace{-5mm}
\end{figure}

Despite their effectiveness, existing search agents rely on large language models (LLMs, typically $\ge$7B parameters). This reliance hinders real-world deployment with latency or budget constraints, where small language models (SLMs, $<4$B parameters) are preferred for their efficiency~\cite{belcak2025smalllanguagemodelsfuture}. This motivates our central research question: \textbf{How can agentic search capabilities be effectively distilled into SLMs?}

We first evaluate the performance of small language models (SLMs; Qwen3-1.7B) as search agents on multi-hop QA benchmarks and identify a central failure mode: \emph{parametric hallucination}. This issue is particularly pronounced in SLMs, which tend to over-rely on limited parametric knowledge and generate speculative answers. Moreover, naively distilling agent trajectories from large language models (LLMs; Qwen3-32B) yields marginal improvements (F1 score from 50.6 to 53.2 in Figure~\ref{fig:bambooge_comaprison}), since LLM-generated trajectories often encode reasoning steps that implicitly depend on parametric knowledge unavailable to SLMs. 

To address parametric hallucination, we propose \textbf{\policy (ASP)}, a distillation paradigm that explicitly constrains search behavior during training. Rather than allowing SLMs to implicitly rely on parametric knowledge, ASP enforces students to always retrieve and ground all essential information with external search before answering. ASP prioritizes evidence-grounded reasoning and discourages speculative inference that over-relies on parametric knowledge. As a result, ASP substantially narrows the performance gap between SLMs and their larger variants, improving F1 from 53.2 to 70.6 and leaving only a 2.5 gap to the teacher.
 
We further investigate whether SLMs can adaptively decide when to search based on their confidence in parametric knowledge. Probing experiments show that SLMs suffer from substantial performance degradation even when self-answering only 5\% most confident queries. SLMs benefit from consistent and enforced retrieval, with ASP emerging as the most effective policy for maximizing SLM search agent performance.

%% file: latex/2_taskformulation.tex
\section{Task Formulation}

We formalize agentic search as a multi-step decision-making process similar to recent work \cite{li2025search}. Given a question $Q$, the agent aims to predict the answer $\hat{y}$ by interacting with an external search tool through the trajectory $\mathcal{T} = (s_1, a_1, o_1, \cdots, s_n, a_n,o_n, \hat{y})$. At each step $t$, the model generates a thought $s_t$ to plan its next move, followed by an action $a_t \in \mathcal{A}$. The action space $\mathcal{A}$ primarily consists of: (\romannumeral 1) \textbf{Search}: formulating a query $q_t$ that invokes the search tool to get the top-$k$ documents from the corpus $\mathcal{D}$. (\romannumeral 2) \textbf{Answer}: terminating reasoning and returning the answer $\hat{y}.$  Following common practice~\cite{li2025search, xu2025reconreasoningcondensationefficient}, we also adopt a LLM summarizer as reason-in-document module to condense output documents into the observation $o_t$.

%% file: latex/3_currentapproach.tex
\section{Evaluating Vanilla and Distilled SLM Agents}

In this section, we evaluate the agentic performance of the Qwen3 series across varying scales with detailed setup in appendix~\ref{app:evaluation_setup} and identify the major bottlenecks with existing standard distillation. 
Section~\ref{sec:zero-shot} and Section~\ref{sec:initial_results} present performance of vanilla and distilled models and further analysis into the failure mode.

\subsection{Vanilla Performance}
\label{sec:zero-shot}

\begin{figure}[h] 
    \centering
     \includegraphics[width=\columnwidth]{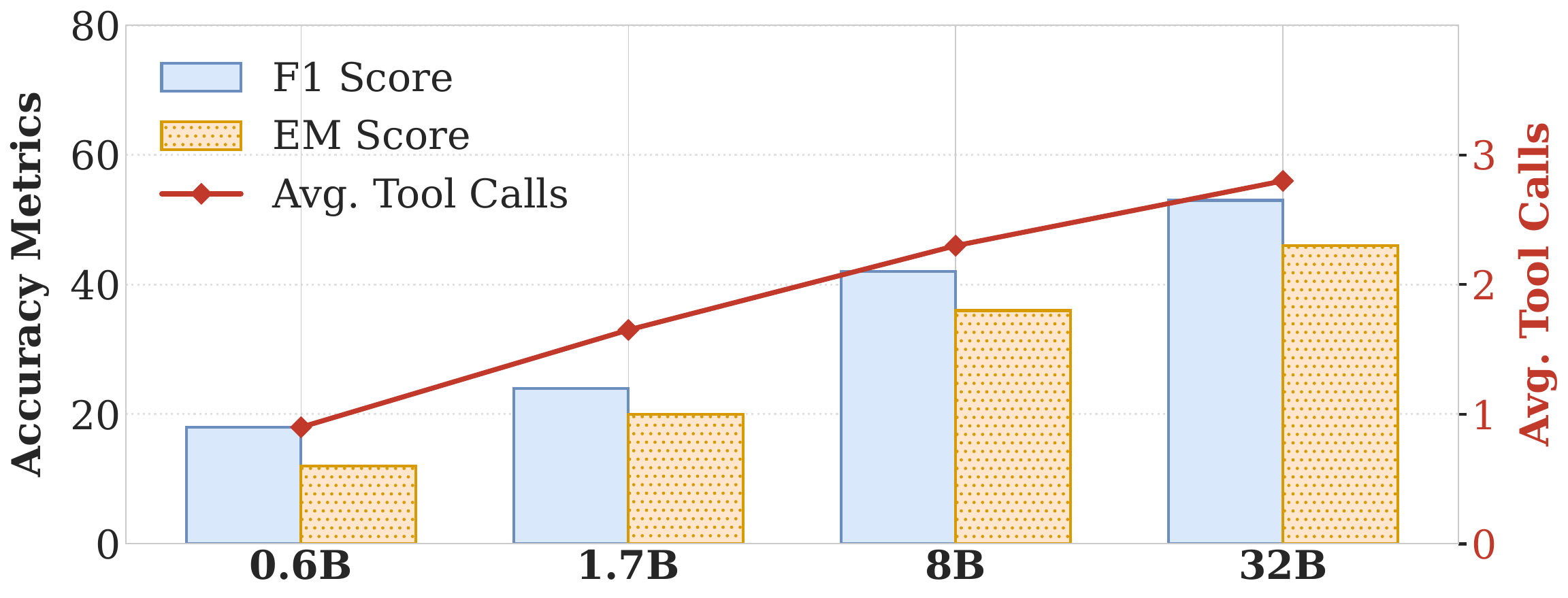}
    \caption{Scaling of agentic search performance. Small models trail behind in both performance and retrieval capability on HotpotQA.}
    \label{fig:scaling}
\end{figure}

As illustrated in Figure~\ref{fig:scaling}, 
compared to their larger variants ($\ge$ 8B), SLMs exhibit significantly worse performance along with fewer tool-calling frequencies. 
The decreased usage of external tools is not intuitive as small models usually have more limited parametric knowledge.
Manual inspection reveals that they either hallucinate answers or struggle with syntactic requirements of tool calling.
The failure to use external tools likely contributes to SLM's ultimate failure in providing a correct final answer.

\subsection{Agentic Distillation} \label{sec:initial_results}
A standard and intuitive approach to improve the agentic capability of SLMs is agent distillation~\citep{kang2025distillingllmagentsmall}. Specifically, given successful trajectories generated by a teacher model, the student model is optimized using standard Supervised Fine-Tuning (SFT) to maximize the probability of the trajectories.
 \paragraph{Setups.} We randomly sample 18,000 questions from the HotpotQA training set for trajectories generation by the teacher model Qwen3-32B and train Qwen3-1.7B with trajectories with String-F1 higher than $0.65$.
After distillation, we evaluate the performance on 500 questions from HotpotQA development set.
Experimental details can be found in Appendix~\ref{app:exp_details}.

\paragraph{Results.} Despite successful convergence, the distilled model still exhibits a substantial performance gap compared to teacher model on evaluation set (47.9\% vs 60.3\%) with 1.89 search tool calls per question compared to 3.02 for the teacher model.
Human annotation reveals that insufficient search and hallucination are the major causes of failure (more details in Appendix~\ref{app:error_ana}).
Note that in standard agentic distillation, while the teacher model is provided with the search tool, it is still allowed to use its own parametric knowledge and often does so.

To further analyze the impact of retrieval quality on downstream performance, we select 926 questions from HotpotQA~\cite{yang2018hotpotqadatasetdiverseexplainable} and 2WikiMultiHopQA~\cite{ho2020constructingmultihopqadataset} that are correctly answered by Qwen3-32B. We extract the retrieved evidence from these successful trajectories and provide it as context to a smaller Qwen3-1.7B model for answer generation. As a result, the 1.7B model’s accuracy improves from 47.9\% to 74.7\%, indicating that the primary performance bottleneck lies in retrieving relevant information rather than in reasoning capability.

%% file: latex/4_search_only.tex
\section{Distillation with Always-Search Policy}
\label{sec:search}

\begin{table*}[h] %
\centering
\small
\captionsetup{justification=justified,singlelinecheck=false}

\scalebox{0.95}{\begin{tabular}{ l c c c c c c c c}
\toprule
\textbf{Models} & \textbf{ASP} &
\textbf{HotpotQA} &
\textbf{2Wiki} &
\textbf{Bamboogle} &
\textbf{MuSiQue} &
\textbf{BrowseComp} &
\textbf{Frames} &
\textbf{LongSeAL} \\
\midrule
Vanilla-Qwen3-0.6B  & \no    & 19.4 & 26.0 & 34.4 & 8.7 & 4.6 & 8.9 & 7.4 \\
Vanilla-Qwen3-1.7B  & \no   & 42.3 & 39.8 & 50.6 & 22.9 & 3.2 & 15.2 & 9.0 \\
Vanilla-Qwen3-4B    &  \no  & 53.4 & 54.7 & 69.1 & 28.7 & 7.0 & 28.4 & 8.2 \\
Vanilla-Qwen3-8B   &  \no   & 58.2 & 58.1 & 71.5 & 32.0 & 16.0 & 30.2 & 10.8 \\
Vanilla-Qwen3-32B  &  \no   & 60.3 & 69.9 & 73.1 & 32.5 & 21.0 & 36.6 & 13.5 \\
Distilled-Qwen3-1.7B  &  \no  & 47.9 & 44.1 & 53.2 & 25.4 & 10.1 & 16.7 & 6.0 \\
Distilled-Llama3.2-1B  &  \no  & 38.2 & 30.9 & 37.9 & 15.9 & 5.1 & 11.4 & 6.0 \\
Distilled-Llama3.2-3B  &  \no  & 47.1 & 46.3 & 53.6 & 20.6 & 3.5 & 16.4 & 9.8 \\
\midrule
SFT-Qwen3-0.6B     &   \yes   & 47.0 & 47.4 & 62.9 & 22.7 & 9.0 & 21.0 & 7.0 \\
SFT-Qwen3-1.7B     &  \yes    & 57.6 & 58.5 & \textbf{70.6} & \textbf{29.2} & \textbf{10.0} & \textbf{25.2} & \textbf{10.1} \\
OPD-Qwen3-1.7B     &   \yes   & 56.2 & \textbf{62.9} & 61.4 & 28.6 & 8.4 & 24.6 & 8.6 \\
Mixed-Qwen3-1.7B    &  \yes   & \textbf{58.2} & 57.8 & 69.4 & 27.2 & 8.8 & 25.0 & 8.3 \\
SFT-Llama-3.2-1B     &   \yes   & 44.5 & 48.2 & 64.6 & 22.3 & 8.2 & 18.6 & 9.3 \\
SFT-Llama-3.2-3B     &   \yes   & 53.0 & 58.8 & 68.4 & 14.4 & 10.0 & 24.8 & 10.1 \\
\bottomrule
\end{tabular}}

\caption{String-F1 scores across agentic-search benchmarks. Small models distilled with \policy have comparable performance with their larger variants. Mixed means start with SFT and then OPD.}
\label{tab:results}    
\end{table*}

Based on findings in previous section, 
we argue that search is mandatory for SLMs to achieve high performance. During training, we adopt a \ours paradigm on SLMs, where models must search for any related information instead of utilizing their own parametric knowledge.

\subsection{Experiment Setup}
\paragraph{Fine-tuning Techniques.} To show the validity of our proposal, we incorporate Always-Search Policy into standard \emph{Supervised Fine-Tuning (SFT)} \cite{kang2025distillingllmagentsmall}.
Specifically, apart from setting a threshold of 0.65 for String-F1, we also apply search tool checking and keyword filtering. Only trajectories where models consistently use search tools to obtain information (instead of generating an answer with words like ``I remember'') are retained for training. 
For \emph{On-Policy Distillation (OPD)}~\cite{rishabh2024onpolicy,lu2025onpolicydistillation}, rather than explicitly filtering trajectories, we insert system prompt asking models to always use search tools and expect that the teacher's log-probability distribution can regulate SLM's behavior and encourage searching.
We also include a \emph{Mixed} setting where OPD is performed on top of ASP-incorporated SFT to further reinforce the behavior.
As a downstream enhancement, we apply \emph{Rejection Fine-Tuning (RFT)} \cite{yuan2023scalingrelationshiplearningmathematical} 
as a final stage to further exploit the capacity of distilled models by selectively reinforcing high-quality agentic behaviors.

\paragraph{Evaluation.} We evaluate Qwen3 and Llama-3.2 family of different model sizes as well as SLM agents trained with various ASP-incorporated distillation methods. Detailed model specifications and experimental configurations are provided in Appendix~\ref{app:exp_details}. We report String-F1 on various agentic search benchmarks: (\romannumeral 1) \textbf{ Structured Multi-hop Reasoning}: \textit{HotpotQA, 2WikiMultiHopQA, Bamboogle and MuSiQue} \cite{yang2018hotpotqadatasetdiverseexplainable, ho2020constructingmultihopqadataset, press2023measuringnarrowingcompositionalitygap, trivedi2022musiquemultihopquestionssinglehop}; (\romannumeral 2) 
\textbf{Agentic \& Information Seeking QA}: \textit{BrowseComp-plus, Frames and LongSeAL}. \cite{chen2025browsecompplusfairtransparentevaluation, krishna2025factfetchreasonunified, pham2025sealqaraisingbarreasoning}.

\subsection{Results}

Table~\ref{tab:results} presents the main results across structured multi-hop reasoning and complex information-seeking benchmarks.

\paragraph{The performance gap narrows after ASP.} Our proposed  \policy significantly enhances the capability of SLMs. Generally, all the three distillation methods with ASP show a comparable performance with Qwen3-8B.
Specifically, in HotpotQA, the 1.7B model trained under mixed setting achieves 58.2 point on the in-distribution test set, matching the 8B model. 
For 2WikiMultihopQA, OPD yields an even higher score 
than Qwen3-8B. Notably, although all training is performed with the training set of HotpotQA, all trained models generalize well to out-of-distribution agentic search tasks, including more challenging benchmarks such as BrowseComp-Plus, Frames, and LongSeAL. 

\paragraph{Consistent search tool calling.} ASP effectively mitigates the ``under-searching'' tendency of SLMs shown in Figure~\ref{fig:scaling}. Compared to the vanilla model (1.72 searches per question), ASP increases the search frequency to 2.47 (SFT) and 2.84 (OPD). This active tool-calling behavior is crucial for reducing hallucination and improving the performance.

\paragraph{Robustness to noisy retrieval.} To evaluate ASP-trained models' performance in noisy retrieval, we replace 10\% retrieval results with failed retrieval. 
Vanilla SLMs and Distilled-1.7B suffer a significant drop (12.1), whereas ASP-trained models exhibit more stable behavior with only 2.3 and 1.7 respectively.
This suggests that ASP also improves the model's ability to recover when retrieval fails.

%% file: latex/5_analysis.tex
\section{Should SLMs Search Adaptively?}

Adaptive search has been proposed as a mechanism to reduce computational overhead by allowing search agents to selectively invoke retrieval tools \cite{eisenstein2025don}. To assess whether this adaptive search is viable for SLMs-based search agents, we introduce a confidence probe to exam a model’s ability to answer with internal knowledge. The probe implementation is detailed in Appendix~\ref{app:probe_setup}. All probes are well calibrated, providing a reliable signal for evaluating \adaptive.

\paragraph{SLMs Know Less.} Appendix~\ref{app:confidence} presents the distribution of confidence on their knowledge to search queries. Teacher model, expected to have more parametric knowledge, holds more than 40\% high confidence queries. In contract, small models' confidence distribution suggests that they lack reliable internal knowledge for the key information.

\paragraph{SLMs Should Always Search.} We simulate adaptive search by filtering queries with top-$P$ confidence and prompting SLMs to answer based on their parametric knowledge.  As shown in table~\ref{tab:threshold}, we see a significant performance difference between model scales. LLMs can effectively self-answer more than 10\% of queries without sacrificing accuracy. In contrast, SLMs exhibit a significant performance drop at a lower rate (e.g., $P=5\%$). Therefore, we argue that the best policy for SLMs is to always search.

\begin{table}[ht!]
\centering
\small
\setlength{\tabcolsep}{4pt} 
\resizebox{\columnwidth}{!}{

\begin{tabular}{@{}lcccc@{}}
\toprule
\textbf{Model} & \textbf{P=1\% $\downarrow$}  & \textbf{P=5\% $\downarrow$}  & \textbf{P=10\% $\downarrow$} & \textbf{P=20\% $\downarrow$} \\ \midrule
Vanilla-Qwen3-32B      & $<$0.1   & 0.3   & 0.6  & 5.2    \\
SFT-Qwen3-1.7B            & 1.9    & 4.8   & 15.0   & 22.0         \\
OPD-Qwen3-1.7B           & 0.8   & 9.0   & 14.5    & 18.0            \\ \bottomrule
\end{tabular}
}
\caption{ After applying adaptive search with different top-$P$ setting on dataset HotpotQA, Vanilla-Qwen3-32B and SFT-Qwen3-1.7B's performance drop}
\label{tab:threshold}
\end{table}

%% file: latex/7_relatedwork.tex
\section{Related Works}
\paragraph{LLM-based Search Agents.} Augmenting language models with external search tools has significantly expanded their abilities for knowledge-intensive tasks~\cite{yao2023reactsynergizingreasoningacting, li2025search, jin2025searchr1trainingllmsreason}. LMs interact with external retrieval tools in reasoning loops to help answer complex, multi-hop questions. However, most search agents are based on LLMs,
whose computational cost and high latency limit their real-world deployment. 

\paragraph{Distillation for Efficient Agents.} Knowledge distillation, \cite{hinton2015distillingknowledgeneuralnetwork} has been adopted to transfer capabilities from teacher model to student model. Chain-of-Thought (CoT) distillation \cite{wei2023chainofthoughtpromptingelicitsreasoning, magister2023teachingsmalllanguagemodels} has improved the reasoning skills of SLMs. Recent works ~\cite{chen2023fireactlanguageagentfinetuning,zeng2023agenttuningenablinggeneralizedagent} explore distilling tool-use capabilities. While these methods show promise, they often retain the nature of the teacher.
Our method differs by enforcing \policy during the distillation process.

%% file: latex/6_conclusion.tex
\section{Conclusion}

In this paper, we explore the performance gap between SLMs and LLMs on complex QA tasks.
Through trajectory analysis, we identify that SLMs' tendency to rely on parametric knowledge rather than invoking search tools is the primary cause of their underperformance, and standard distillation approach fails to adequately address the issue.
To remedy this, we propose ASP into the distillation process and demonstrate its effectiveness when combined with distillation methods.
Our results show that by enforcing search tool usage, SLMs can achieve performance comparable to LLMs across benchmarks.
Further analysis through confidence probing validates that ASP is necessity for SLMs.
Our work examines the failure modes of SLMs on complex QA tasks and underscores the importance of prioritizing tool usage for SLMs.

\section*{Limitations}
While \policy offers a paradigm on how to train SLM agents to achieve stronger performance, our current training strategy adopts a relatively simple instantiation of this paradigm. How to integrate \policy into more advanced training frameworks to further unlock the potential remains an open direction for future work. 

Meanwhile, we do not rigorously characterize the upper bound of SLM-based agents, which is influenced by multiple factors beyond retrieval behavior, such as reasoning capacity. Systematically exploring these factors would provide a deeper understanding of the limits and opportunities of SLM-based agents. 

In addition, the effectiveness of \policy implicitly assumes that retrieved information is always accurate and reliable, whereas real-world search environments often contain noisy or misleading content. Developing mechanisms to robustly handle such noise is another crucial aspect. 

Finally, our evaluation focuses on the Qwen3 model family. Validating our findings across a broader range of LM architectures would further strengthen our proposed approach.

%% file: latex/8_appendix.tex
\clearpage

\section{Algorithms}
\subsection{Algorithm for Agent Distillation}
\begin{algorithm}[h]
    \caption{Agentic Search}
    \label{alg:search_agent}
    \begin{algorithmic}[1]
        \Require Query $x$, LMs $\pi_\theta$, Search Tool $\mathcal{S}$, Max Hops $\mathcal{H}$
        \Ensure Answer $y$
        
        \State $\tau \leftarrow [x]$
        \For{$t = 1, \dots, \mathcal{H}$} \Comment{Max turn to avoid infinite loop}
            \State $s_{t} \leftarrow \pi_\theta(\tau)$ \Comment{Sample response from LLM}
            
            \If{$r$ contains \texttt{<search>}$a_t$\texttt{</search>}}
                \State Parse query $a_t$ from $s_t$
                \State $o_t \leftarrow \mathcal{S}(a_t)$ \Comment{Execute search}
                \State $\tau \leftarrow \tau \oplus s_t \oplus \texttt{<information>} o_t \texttt{</information>}$ \Comment{Append observation}
            
            \ElsIf{$s_t$ contains \texttt{<answer>}$y$\texttt{</answer>}}
                \State \Return $y$ \Comment{Finish if task complete early}
            \Else
                \State $\tau \leftarrow \tau \oplus s_t$ \Comment{Chain-of-thought reasoning}
            \EndIf
        \EndFor
        
        \State \Return $y$ \Comment{Fallback max hops reached}
    \end{algorithmic}
\end{algorithm}

\subsection{Algorithm for \ours Distillation}
\begin{algorithm}[h]
    \caption{\ours}
    \label{alg:pf_search}
    \begin{algorithmic}[1]
        \Require Query $x$, Policy $\pi_\theta$, Tool $\mathcal{S}$, Constraint Prompt $\mathcal{P}_{\text{force}}$, Max Retries $K$
        \Ensure Answer $y$ and Valid Trajectory $\tau$ or Error
        
        \State $\tau \leftarrow [x \oplus \mathcal{P}_{\text{force}}]$ \Comment{Force model to rely on tools, inject constraint globally at the start}
        
        \For{$t = 1, \dots, \mathcal{H}$}
            \State $k \leftarrow 0$
            \Repeat
                \State $s_{t} \leftarrow \pi_\theta(\tau)$ \Comment{Sample response with action or answer}
                \State $k \leftarrow k + 1$
            \Until{\big(HasAction($s_t$) \textbf{or} IsAnswer($s_t$)\big) \textbf{or} $k \geq K$}
            
            \If{\textbf{not} (HasAction($s_t$) \textbf{or} IsAnswer($s_t$))}
                \State \textbf{Raise Error} (Discard Trajectory) \Comment{Failed to enforce search}
            \EndIf

            \If{$s_t$ contains \texttt{<search>}$q_t$\texttt{</search>}}
                \State $o_t \leftarrow \mathcal{S}(q_t)$
                \State $\tau \leftarrow \tau \oplus s_t \oplus \texttt{<information>} o_t \texttt{</information>}$
                
            \ElsIf{$s_t$ contains \texttt{<answer>}$y$\texttt{</answer>}}
                \State \Return $y$, $\tau$ \Comment{Successful answer}
            \EndIf
        \EndFor
        
        \State \Return Failure \Comment{Max hops reached}
    \end{algorithmic}
\end{algorithm}

\section{Experimental Details}
\label{app:exp_details}
\subsection{Evaluation Setup}\label{app:evaluation_setup}
We evaluate the agentic performance of the Qwen3 series across varying scales (0.6B to 32B) \cite{qwen3} on \textit{HotpotQA} and \textit{2WikiMultiHopQA} benchmarks~\cite{yang2018hotpotqadatasetdiverseexplainable, ho2020constructingmultihopqadataset}.
We use Qwen3-32B as summarizer and e5-large-v2 retriever \cite{wang2022text} on fullwiki-20210620 corpus \cite{wikipedia_dump_20210620}.
For all tasks we report String-F1 and/or Exact Match (EM) scores between the predicted answer $\hat{y}$ and closest golden answer $y^*$.
\subsection{Training Details}
\paragraph{Agent Distillation:} We sample 18000 trajectories from teacher model based on questions from training set of HotpotQA benchmark and filter out trajectories that lead to wrong answer. The distillation is performed over 3.0 epochs using the AdamW optimizer. We set a learning rate of $1e-5$ and a batch size of 4. 
\paragraph{Trajectory-based Offline Distillation:} The settings are same to agent distillation except that we control \ours behavior by prompting and strict filtering on the trajectories.
\paragraph{On-Policy Distillation:} We train the model on 3000 samples from HotpotQA training set. For each question, we sample 8 trajectories from the student model. With a batch of 4 questions, the trajectories go into the teacher model for token probability distribution.  We optimize the Kullback-Leibler (KL) Divergence loss in batch. The distillation is performed over 4.0 epochs using the AdamW optimizer. We set a learning rate of $2e-6$. The student model is prompted to follow the \policy and the teacher model will observe and restrain the action of smaller model. 
\paragraph{Rejection Fine-tuning:} We use 10000 trajectories generated by the student model based on questions from training set of HotpotQA benchmark (different questions to agent distillation). The distillation is performed over 2.0 epochs using the AdamW optimizer. We set a learning rate of $5e-6$ and a batch size of 4.

\subsection{Experiment Configurations}
\paragraph{Retriever and Corpus:} On BrowseComp-Plus, we use embedding-based retrieval with Qwen3-Embedding-8B as the embedding model and the corpus provided in the BrowseComp-Plus benchmark. On all other benchmarks, we use e5-large-v2 retriever on fullwiki-20210620 corpus. 
\paragraph{Summarizer:} We use Qwen3-32B as summarizer. The model is set to non-thinking mode with Temperature=0.7, TopP=0.95, TopK=20, and MinP=0. 

\subsection{Prompts}
\begin{promptbox}{System instruction used for Agentic Search}
    \#\#\# Instruction
    
    Answer the following question from the user with the help of a Wikipedia search engine. Please reason step by step. You should think about what you need to know in order to answer the question, and then search for that information using the search engine. To perform a search operation, write a web search question and enclose it with <search> and </search>. You will immediately observe a piece of summarized search results within the <information> and </information> tags. You can then use this retrieved information to continue your reasoning. You can repeat the search process many times. Once you think you have all the information you need, you can end the thinking process and provide the final answer. You MUST enclose your final answer with <answer> and </answer>. 

    \#\#\# Example
    
    Question: When did the people who captured Malakoff come to the region where Philipsburg is located?

    Your thinking process:
    
    Alright, I need to figure out when the people who captured Malakoff came to the region where Philipsburg is located. I should first find out who were the people that captured Malakoff. Let me write a search question to look it up with the Wikipedia search engine.
  
    <search> Who were the people that captured Malakoff? </search>
    
    <information> The French army under General MacMahon successfully captured the Malakoff redoubt on 8th. </information>
  
    Okay, so the French people captured Malakoff. Now, the next step would be to figure out in what region Pilipsburg is located. I will write a web search to look that up.
  
    <search> Where is Philipsburg located at? </search>
    
    <information> Philipsburg is is the main town and capital of Sint Maarten, a constituent country of the Kingdom of the Netherlands. </information>
  
    ...[more thoughts shortened]...

    Your final response:
    
    <answer> November 12, 1625 </answer>

    \#\#\# Reminders
    
    1. You should carefully follow the format of searching and answering as shown in the example above.
    
    2. You should always and directly use the Wikipedia search engine to look up the information needed to answer the question.
    
    3. Your search queries should be a complete, natural language question instead of keywords. For instance, instead of searching for "people that captured Malakoff", you should search for "Who were the people that captured Malakoff?".
    
    4. You final answer should be a short-form answer. Do NOT provide explanations or extract descriptions. For instance, instead of saying "The people who captured Malakoff came to the region where Philipsburg is located in November 12, 1625", you should only say "November 12, 1625".
    
    5. Your final response should only have the answer enclosed in <answer> and </answer> tags. Do not include any other information or text.
    
    6. Assume that the current year is 2018. No need to look up information that is more recent than this year.

    \#\#\# Search Query Format Guidelines
    
    When writing search queries, follow these specific formats depending on what information you need:

    **Format A: When inquiring about an attribute of an entity**
    
    Use the pattern: "wh-word is [attribute] of [entity]"
    
    Examples:
    
    - "What is the capital of France?" (asking about the capital attribute of France entity)
    
    - "When is the birthday of Albert Einstein?" (asking about the birthday attribute of Albert Einstein entity)
    
    - "Where is the location of Mount Everest?" (asking about the location attribute of Mount Everest entity)

    **Format B: When inquiring about entities that satisfy given values on given attributes**
    
    Use the pattern: "wh-word [satisfies the values on given attributes]"
    
    Examples:
    
     - "Who was born in 1879?" (asking about entities with birth year attribute = 1879)
     
    - "Which countries have a population over 100 million?" (asking about entities with population attribute > 100 million)
    
    - "What cities are located in California?" (asking about entities with location attribute = California)

    Choose the appropriate format based on whether you're looking for an attribute value of a specific entity (Format A) or searching for entities that match certain criteria (Format B).
  
    Now let's begin!

    \#\#\# Question
    
    \{question\}
\end{promptbox}

\begin{promptbox}{System instruction for \ours}
    \#\#\# Instruction

    \textbf{You are a "Knowledge-Free" agent. You are not allowed to use any of your internal pre-trained memory or knowledge base. You must act as if you know nothing about the world.} \textbf{You MUST use the search engine to verify EVERY single entity and fact mentioned in the question, even if it seems like common sense.} Please reason step by step. To perform a search operation, write a web search question and enclose it with <search> and </search>. You will immediately observe a piece of summarized search results within the <information> and </information> tags. You can then use this retrieved information to continue your reasoning. Once you think you have all the information you need, you can end the thinking process and provide the final answer. You MUST enclose your final answer with <answer> and </answer>. \textbf{Any reasoning step NOT supported by a search result will be considered a hallucination.}

    \#\#\# Example
    
    Question: When did the people who captured Malakoff come to the region where Philipsburg is located?

    Your thinking process:
    
    Alright, I need to figure out when the people who captured Malakoff came to the region where Philipsburg is located. I should first find out who were the people that captured Malakoff. Let me write a search question to look it up with the Wikipedia search engine.
  
    <search> Who were the people that captured Malakoff? </search>
    
    <information> The French army under General MacMahon successfully captured the Malakoff redoubt on 8th. </information>
  
    Okay, so the French people captured Malakoff. Now, the next step would be to figure out in what region Pilipsburg is located. I will write a web search to look that up.
  
    <search> Where is Philipsburg located at? </search>
    
    <information> Philipsburg is is the main town and capital of Sint Maarten, a constituent country of the Kingdom of the Netherlands. </information>
  
    ...[more thoughts shortened]...

    Your final response:
    
    <answer> November 12, 1625 </answer>

    \#\#\# Reminders
    
    1. You should carefully follow the format of searching and answering as shown in the example above.
    
    2. You should always and directly use the Wikipedia search engine to look up the information needed to answer the question. \textbf{Do not use your own knowledge or personal experiences to speculate.}
    
    3. Your search queries should be a complete, natural language question instead of keywords. For instance, instead of searching for "people that captured Malakoff", you should search for "Who were the people that captured Malakoff?".
    
    \textbf{4. You can trust the information retrieved from the search engine to be accurate and factual, and use it in your subsequent reasoning. No need to reflect and doubt it.}
    
    5. You final answer should be a short-form answer. Do NOT provide explanations or extract descriptions. For instance, instead of saying "The people who captured Malakoff came to the region where Philipsburg is located in November 12, 1625", you should only say "November 12, 1625".
    
    6. Your final response should only have the answer enclosed in <answer> and </answer> tags. Do not include any other information or text.
    
    7. Assume that the current year is 2018. No need to look up information that is more recent than this year.

    \#\#\# Search Query Format Guidelines
    
    When writing search queries, follow these specific formats depending on what information you need:

    **Format A: When inquiring about an attribute of an entity**
    
    Use the pattern: "wh-word is [attribute] of [entity]"
    
    Examples:
    
    - "What is the capital of France?" (asking about the capital attribute of France entity)
    
    - "When is the birthday of Albert Einstein?" (asking about the birthday attribute of Albert Einstein entity)
    
    - "Where is the location of Mount Everest?" (asking about the location attribute of Mount Everest entity)

    **Format B: When inquiring about entities that satisfy given values on given attributes**
    
    Use the pattern: "wh-word [satisfies the values on given attributes]"
    
    Examples:
    
     - "Who was born in 1879?" (asking about entities with birth year attribute = 1879)
     
    - "Which countries have a population over 100 million?" (asking about entities with population attribute > 100 million)
    
    - "What cities are located in California?" (asking about entities with location attribute = California)

    Choose the appropriate format based on whether you're looking for an attribute value of a specific entity (Format A) or searching for entities that match certain criteria (Format B).
  
    Now let's begin!

    \#\#\# Question
    
    \{question\}
\end{promptbox}

\begin{promptbox}{System instruction for Summarizers}
    \#\#\# Task
    
    You are given a user query and a set of retrieved documents. Your job is to extract a concise, factual, and relevant answer to the query, using only information from the provided documents.

    \#\#\# Instructions
    
    1. Carefully read each document and determine if it contains information relevant to the query.
    
    2. If you find relevant information, extract and summarize it in 1-3 clear sentences.
    
    3. **Do not use any information that is not present in the documents.**
    
    4. If none of the documents contain relevant information, state that clearly.

    \#\#\# Output Format (CRITICAL - MUST FOLLOW EXACTLY)
    
    - Your answer **MUST start with exactly**: \#\#\# Extracted Information
    
    - On the line(s) after this tag, write the extracted information.
    
    - If there is no relevant information, write: No helpful information found.
    
    - **IMPORTANT**: Even if the documents are long, you MUST start your answer with \#\#\# Extracted Information

    \#\#\# Example Output

    \#\#\# Extracted Information
    
    [Your extracted answer here.]

    or

    \#\#\# Extracted Information
    
    No helpful information found.

    \#\#\# User Query
    
    \{question\}

    \#\#\# Documents
    
    \{documents\}
\end{promptbox}

\section{Error analysis}
\label{app:error_ana}
We analyze 66 case where the teacher model answers correctly while agent-distilled 1.7B model fails by a human-based categorization on the primary causes of errors.
\textbf{(\romannumeral 1) Insufficient/Bad Retrieval (23/66):} Insufficient retrieval of external information;
\textbf{(\romannumeral 2) Hallucination (33/66):} False self-knowledge use or distrust of retrieved information; 
\textbf{(\romannumeral 3) Reasoning Error (13/66):} Wrong problem understanding or wrong answer based on sufficient information; 
\textbf{(\romannumeral 4) Others (13/66)}. 
Among them, we witness both insufficient search and hallucination in 16 cases, suggesting the high correlation of both issues.

\section{Case Study}
To qualitatively evaluate the performance of distillation on SLM, we conduct a detailed case study on a multi-hop question.

\begin{center}
\small
\resizebox{\columnwidth}{!}{
\noindent\begin{tabularx}{\linewidth}{@{} l L @{}}
    \toprule
    \textbf{Context} & \textit{Question: What is the birth country of the director of the film "The Great Silence"} \\
    \midrule
    \textit{\textbf{Vanilla-Qwen3-32B}} & \texttt{<think>OK, to answer the question, I first need to know who is the director of the file "The Great Silence"</think><search>Who is the director of the film "The Great Silence"<search><information>Sergio Corbucci was the director of "The Great Silence"</information><think>Then, I need to check the nationality of Sergio Corbucci<think><sesarch>Where was Sergio Corbucci born</search><information>Sergio Corbucci was born in Italy, 6 December 1926<information><answer>Italy</answer>} \\
    \midrule
    \textit{\textbf{Vanilla-Qwen3-1.7B}}  & \texttt{OK, I need to search for the director of  "The Great Silence". \textbf{(no search tool calling)} Wait, I remember the film was directed by Sergio Leone, let me check it again. \textbf{doubting loop} Sergio Corbucci sounds like Spanish. \textbf{(hallucination)}<answer>Spain</answer>} \\
    \midrule
    \textit{\textbf{SFT}}  & \texttt{<think>I need to first confirm the director of the film.</think><search>Who is the director of "The Great Silence"</search><information>Sergio Corbucci directed "The Great Silence"</information><think>To get the answer, I need to search of the nationality.</think><search>Sergio Corbucci's nationality.</search><information>Sergio Corbucci was an Italian film director.</information><think>The information is sufficient to conclude.</think><answer>Italy</answer>} \\
    \bottomrule
\end{tabularx}
}
\end{center}

\textbf{Formalizing Tool-Calling Syntax.} One of the primary barriers for SLM is the inability to adhere to structural constraints. As discussed in section~\ref{sec:initial_results}, even when prompted with examples of schemas, model always leaks into natual language, making external tool calling undetectable.

\textbf{\policy} With \policy, SLMs are able to access sufficient information as the scenario described in section~\ref{sec:initial_results}. Therefore, rather than ``guessing'' for an answer, the model can conclude precisely from the collected information.

\textbf{Resolving Reasoning Inertia.} We further compare the untrained model with distilled ones. We found that even when it is able to invoke tools, the model often suffer from doubting the retrieved information. In contrast, the distilled models perform zero doubting and continue the reasoning.
\newpage

\section{Probe Experiment Setup} \label{app:probe_setup}
\subsection{Probe Architecture and Hyperparameter Settings}
To capture the internal uncertainty of the Small Language Models (SLMs), we train a probing classifier on the fixed representations of the backbone model. Let $L$ denote the total number of layers in the SLM. We extract the hidden states from the last four layers, denoted as $\{h_{L-3}, h_{L-2}, h_{L-1}, h_L\}$.

The input to the probe, $x_{probe}$, is constructed by concatenating these hidden representations:
\[x_{probe} = \text{Concat}(h_{L-3}, h_{L-2}, h_{L-1}, h_L)\]

The probe is implemented as a three-layer Multi-Layer Perceptron (MLP) with a strictly decreasing hidden dimension size. The architecture follows a structure of Linear $\rightarrow$ ReLU $\rightarrow$ Linear, specifically mapping the dimensions as $d_{model} \times 4 \rightarrow 512 \rightarrow 256 \rightarrow 128 \rightarrow 1$ (logit).

\begin{table}[h]
    \centering
    \vspace{2mm} %
    \resizebox{\columnwidth}{!}{
    \begin{tabular}{ll}
        \toprule
        \textbf{Configuration} & \textbf{Value / Setting} \\
        \midrule
        \multicolumn{2}{l}{\textit{Architecture Details}} \\
        Probe Type & Multi-Layer Perceptron (MLP) \\
        Number of Layers & 3 \\
        Hidden Dimensions & \{512, 256, 128\} \\
        Activation Function & ReLU \\
        Input Features & $(h_{L-3}, h_{L-2}, h_{L-1}, h_L)$ \\
        \midrule
        \multicolumn{2}{l}{\textit{Training Setup}} \\
        Optimizer & AdamW \\  %
        Learning Rate & $2e-6$ \\ %
        Batch Size & 16 \\ %
        Backbone Status & Frozen \\
        \bottomrule
    \end{tabular}
    }
    \caption{\textbf{Hyperparameters and architectural details for the Confidence Probe.} The probe utilizes features from the final stages of the backbone model to predict generation confidence.}
    \label{tab:probe_hyperparams}
\end{table}

\subsection{Calibration Statistics}

Table \ref{tab:probe_hyperparams} summarizes the specific hyperparameters used for training the probes. We freeze the backbone SLM parameters during the probe training to ensure that the probe reflects the intrinsic knowledge of the pre-trained model without altering its original behavior.

As mentioned, ensuring the probes are well-calibrated is crucial for the validity of our search-saving strategy. Table \ref{tab:ece_stats} presents the detailed Estimated Calibration Error (ECE) and Accuracy on the validation set

\begin{table}[h]
    \centering
    \vspace{2mm}
    \small
    \begin{tabular}{lcc}
        \toprule
        \textbf{Model Name} & \textbf{ECE Score} & \textbf{Accuracy} \\
        \midrule
        Vanilla-Qwen3-32B & 0.041 & 93.7\% \\
        Vanilla-Qwen3-1.7B & 0.052 & 89.6\% \\
        SFT-Qwen3-1.7B   & 0.034 & 95.1\% \\
        OPD-Qwen3-1.7B & 0.045 & 97.3\% \\
        \midrule
        \textbf{Average} & $\mathbf{0.043 \pm 0.009}$ & $\mathbf{93.9\% \pm 3.4\%}$\\
        \bottomrule
    \end{tabular}
    \caption{\textbf{Calibration Performance.} The Estimated Calibration Error (ECE) for the confidence probes across different LMs. Lower is better. The Accuracy on the validation data across different LMs. Higher the better}
    \label{tab:ece_stats}
\end{table}

\begin{figure*}[t]
    \centering
    \includegraphics[width=\linewidth]{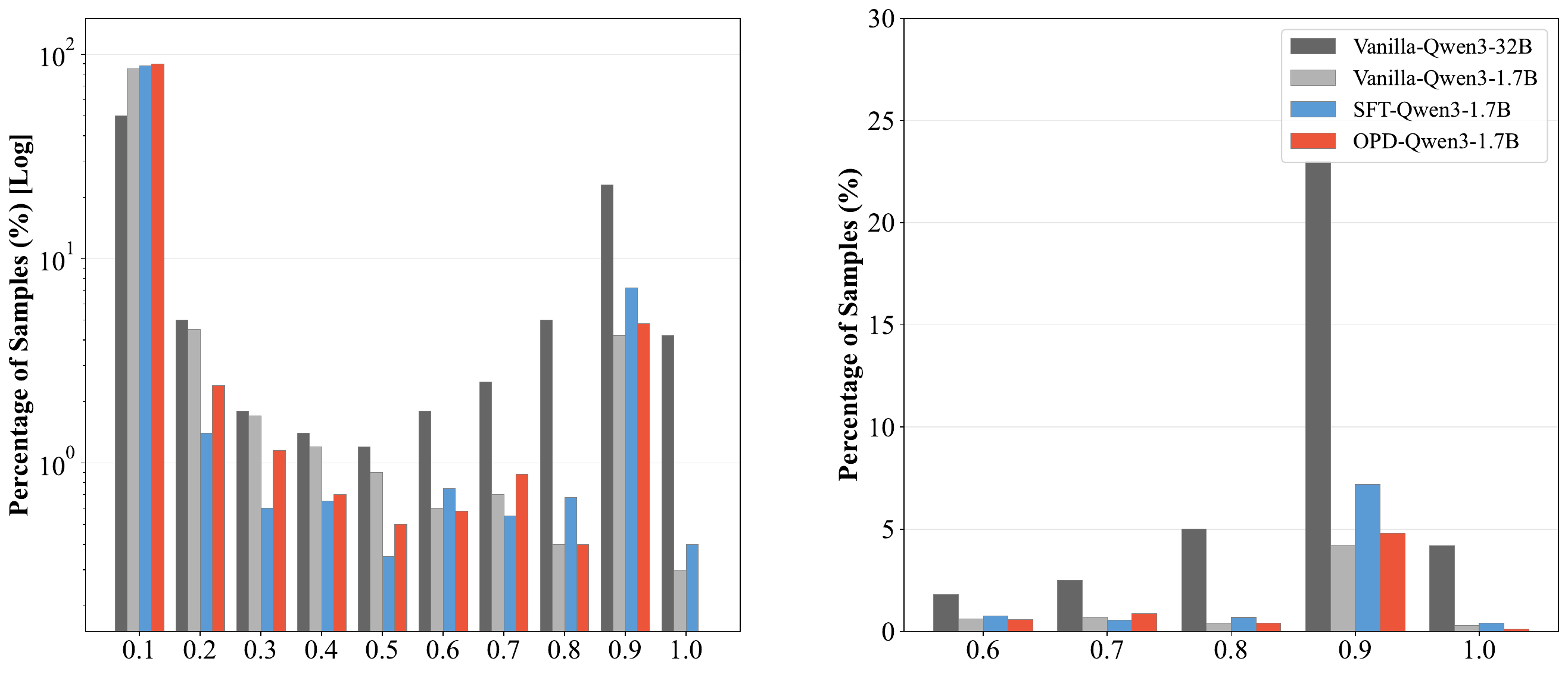}
    \caption{Confidence probing results. (a) illustrates the full sample distribution on a log scale; (b) zooms into high-confidence bins (0.5--1.0).}
    \label{appendix:probing_results}
\end{figure*}

\section{Confidence Distribution of different models} \label{app:confidence}
Figure~\ref{appendix:probing_results} shows the confidence distribution on whether the models can answer the query based on probing.

\section{Latency in Analysis}
\begin{table}[h]
\centering
\small
\begin{tabular}{l c c}
\toprule
\textbf{Model} & \textbf{Tool Calls} & \textbf{Avg. Latency} \\
\midrule
Vanilla-Qwen3-1.7B   & 1.72 & $\sim$1.8s \\
Distilled-Qwen3-1.7B & 1.89 & $\sim$2.7s \\
SFT-Qwen3-1.7B       & 2.47 & $\sim$3.1s \\
Vanilla-Qwen3-8B     & 2.84 & $\sim$5.6s \\
Vanilla-Qwen3-32B    & 3.02 & $\sim$10.3s \\
\bottomrule
\end{tabular}
\caption{End-to-end inference latency on HotpotQA. Setting: 4$\times$ H20, vLLM, FAISS-GPU. ASP-trained SLMs achieve comparable performance to Qwen3-32B at $\sim$3$\times$ lower latency.}
\label{tab:latency}
\end{table}
As table~\ref{tab:latency} illustrated, with more number of search tool calling on average, the inference latency increases. However, it is still faster than the larger LMs.

\section{Noise Ablation}
\subsection{Experiment setup}
In the noise ablation experiment, we apply two different types of noise to the system. (1) Purely unrelated noise which cause the summarizer to generate ``No useful information found''. (2) Similar but unrelated noise which might cuase the summarizer to hallucinate and give wrong information.
\subsection{Results}
By analyzing the results, we found that with purely noise, both SFT/OPD trained model will be able to recover from the noise with adaptively re-plan the search strategy, while the vanilla model fails to recover and hallucinate the answer.
For the summarizer hallucinated ones, we also found a similar pattern that both SFT/OPD trained model gain the ability of information validation from the teacher and drop slightly with only 4.8 point compared to 15.3 and 18.7 points respectively for the vanilla model and distilled model.

\section{Reasoning Ability}
\subsection{General Reasoning ability for vanilla models}
\begin{table}[h]
\centering
\small
\begin{tabular}{l c c c}
\toprule
\textbf{Model} & \textbf{MMLU} & \textbf{GSM8K} & \textbf{GPQA} \\
\midrule
Qwen3-32B     & 83.61 & 93.40 & 49.49\\
Qwen3-8B      & 76.89 & 89.84 & 44.44 \\
Qwen3-4B      & 72.99 & 87.79 & 36.87 \\
Qwen3-1.7B    & 62.63 & 75.44 & 28.28 \\
Qwen3-0.6B    & 52.81 & 59.59 & 26.77 \\
SFT-Qwen3-1.7B& 73.28 & 85.34 & 41.26 \\
OPD-Qwen3-1.7B& 74.31 & 86.28 & 41.37 \\
\bottomrule
\end{tabular}
\caption{Reasoning ability of different models on general reasoning tasks.Data collected from Qwen3 Technical Report}
\label{tab:reasoning}
\end{table}
Table~\ref{tab:reasoning} illustrate the models' reasoning ability on general reasoning tasks. Despite the huge gap between the LLMs and SLMs, the SLMs are achieving high enough scores.

In agentic search settings, the reasoning is not complex as the general reasoning tasks which require math calculation and intensive knowledge reasoning. Instead, the reasoning ability are about acquiring information retrieved from corpus and plan for next move. 

\subsection{Reasoning ability acquired from ASP training}
Table~\ref{tab:reasoning} also include the testing results from our ASP trained models. We found that besides the increment in agentic search tasks, the student model also learn reasoning ability from the teacher.